# Handwriting-based Automated Assessment and Grading of Degree of Handedness: A Pilot Study

Smriti Bala, Venugopalan Y. Vishnu, Deepak Joshi

*Abstract*— **Hand preference and degree of handedness (DoH) are two different aspects of human behavior which are often confused to be one. DoH is a person's inherent capability, which is affected by nature and nurture. In this study, we used dominant and non-dominant handwriting traits to measure DoH for the first time, on 43 subjects of three categories- Unidextrous, Partially-Unidextrous, and Ambidextrous. Features extracted from the segmented handwriting signals called strokes were used for DoH quantification. Davies Bouldin Index, Multilayer perceptron, and Convolutional Neural Network (CNN) were used for automated grading of DoH. The outcomes of these methods were compared with the widely used DoH assessment questionnaires from Edinburgh Inventory (EI). The CNN based automated grading outperformed other computational methods with an average classification accuracy of 95.06±3.08% under stratified 10-fold cross-validation. The leave-one-subject out strategy on this CNN resulted in test individual's DoH score which was converted into a 4-point score. Around 90% of the obtained scores from all the implemented computational methods were found to be in accordance with the EI scores under 95% confidence interval. Automated grading of DoH presented in this study could be used in multiple applications concerned with neuroscience, rehabilitation, physiology, psychometry, behavioral sciences, and forensics.**

## I. INTRODUCTION

ALTHOUGH there has been a demarcation between left-handers and right-handers, humans do possess some form of degree of handedness. The degree of handedness is a measurement of someone's capacity to utilize one hand over the other, as opposed to the direction of handedness or hand preference. It is not a binary trait, meaning there is no clear dividing line between left-handedness and right-handedness. Instead, it is a spectrum, with some people being more strongly left-handed or right-handed than others. Degree of handedness is one of the most interesting traits in humans which links its roots to many genetic and environmental factors [1] such as neuroanatomy [2], socio-cultural aspects [3] and preference [3].

Degree of handedness possesses an abundance of behavioral and health informatics which has aided in critical assessments in previous works. It was investigated along with fMRI and EEG recordings for assessing neurodevelopmental disorders, like autism [4] and dyslexia [5] *(which hinder learning since early age)*, to indicate the prevalence of mixed handedness and poorer language prognosis in these groups. Its association with speech and language lateralization has been repeatedly researched [6]. In disorders such as schizotypy and schizophrenia [7], it's often examined as one of the indicative markers, where a previous study found forced writing hand conversion to cause schizotypy [8] due to subsequent changes in handedness. Furthermore, in epileptic individuals, atypical language dominance was found to increase linearly with degree of left-handedness [9].

DoH and interhemispheric asymmetry can also pose grave challenges in controlling brain-computer interface (BCI) innovations which need brain activity for motor imagery, but a recent study showed that prior knowledge of degree of handedness can improve BCI performance [10]. DoH can help develop effective stroke rehabilitation techniques for dominant and non-dominant hand deficits [11][12].

Other than these, degree of handedness has been investigated for psychometric analyses as well. Personality assessments like authoritarianism, prejudice, disgust, and seeking sensations were studied using degree of handedness [13], [14]. A previous study showed traits of anxiety in consistent right-handers and inconsistent left-handers [15]. DoH has also been studied as a predictive marker of cognitive performance [16]. Nevertheless, degree of handedness had been crucial in criminal investigation, crimes related to handwriting forgery, feigning etc., to find the level of handedness[17].

Despite the various and significant uses of assessing it, grading instruments still rely on questionnaires and qualitative assessments. The past and current methods associated with degree of handedness detection widely follow questionnaires from Edinburgh handedness inventory [18]. PET-based measurements [19], fMRI-based measurements [2], EEG-based measurements of cortical excitability [20], are some of the other methods. The methods consisting of neuroanatomical study are very expensive and time-consuming while the questionnaire-based methods are quick but more qualitative than being quantitative. Alternatives such as peg-board test [21], a speed dependent motor test, may not give proper resolutions to outcomes in healthy persons who actively use both hands. Of the many possible measurable activities such as dexterity, handwriting, darts, tapping, ratchet, strength, and endurance, handwriting has been considered the best method to assess DoH [22] for it being highly complex and elaborate fine motor skill.

Therefore, in this study, we have proposed the use of dynamic kinematic features of handwriting along with its static features for degree of handedness assessment. Previously handwriting as an assessment tool has been successfully applied in identifying – (a) stress levels and emotional states [23], [24], (b) diseases such as Alzheimer [25], [26], Parkinson [27]–[29], (c) children's dysgraphia [30], [31], and psychometry [32] (e) personality traits [13]–[15] etc.

A previous study has also reported the differences that arose in functional neuroanatomy of handwriting due to the effect of hand conversion [19]. In fact, handwriting is termed as a complicated motor-cognitive process [33]. It reflects not only our brain performance in different ways but also the brain-body functional linkages and could also identify any sensorimotor dysfunction. Hence, in this study, degree of handedness was measured using handwriting which contributes immensely towards human hand dexterity. The major contributions of this work include the following:

- A novel task of assessment of degree of handedness from handwriting, which is effective, time bound, and economic, is presented. The provided technique is capable of differentiating between individuals based on degree of handedness.
- Use of segments/strokes of handwriting to extract static and dynamic features is presented.
- A systematic grading technique after feature extraction is tested with statistical-(model-less) and established machine learning classification methods-(model-based). Additionally, a multilayered perceptron is trained on these features to generate scores for the same.
- An automated feature extraction method using convolutional neural networks *(CNN)* which can operate on continuous handwriting data is also presented.
- The systematic comparison of estimated degree of handedness scores from the above mentioned methods with the widely used Edinburgh Inventory scores for the individuals is presented.
- The effects of task and features on degree of handedness scores is also discussed.

## II. METHODS

### A. Participants

Twenty healthy male adults (mean ± SD, age: 26 ± 4.6 years, nine left-handed and eleven right-handed), and twenty-one healthy female adults (mean ± SD, age: 25 .2± 5.7 years, eleven left-handed and ten right-handed) participated in this study. Additionally, two ambidextrous male who identified themselves as left-handed (mean ± SD, age: 27 ± 1.5 years) participated in this study. These participants had no neurological, brain injuries, muscular injuries (including bone fractures and any orthopedic surgery in the past), sensory impairment, or related disabilities as per their health history and personal understanding. Ethical clearance for these experiments was approved by the All India Institute of Medical Sciences ethics committee (Ref. No. IEC-299/07.05.2021). Consent of voluntary participation in this study was taken from each participant.

### B. Groups

A preliminary questionnaire was prepared after studying Oldfield Handedness Questionnaire [18], and Annett Questionnaire [34]. It consisted of all 10 items of Edinburgh Inventory [18]. This was done to assess preliminary information about subject's position in between the two extreme ends of being purely "Unidextrous" or "Ambidextrous". Calculations of handedness score were done according to Edinburgh Inventory. The subjects comprised many first timers who never wrote with their non-dominant hand, some "in-practice" persons who used their non-dominant hand for writing due to personal choice, some sports people who preferred their non-writing hand for playing games, some converted left-handers who practiced writing with right hand and gross motor activities with left hand, and ambidextrous subjects. Three groups were formed which were named as "Unidextrous" (U) consisting of pure left handers and right handers who were first-timers, "Partially-Unidextrous" (PU) consisting of left-handers and right-handers who frequently trained their non-dominant hand with writing activities, sports, and other chores, and "Ambidextrous" (A) consisting of two males who could write with both hands with equal fluency. The group (PU) included active users of non-dominant hand, whether due to sports, personal preference, or conversion. The group (A) had two male subjects, one (Subject ID S22) had a condition similar to hematidrosis, preventing him to use one hand for long hours and the other (Subject ID S25) used both hands even after conversion. The subjects themselves identified their dominant and non-dominant hand as well as their respective group based on their writing hand preference.

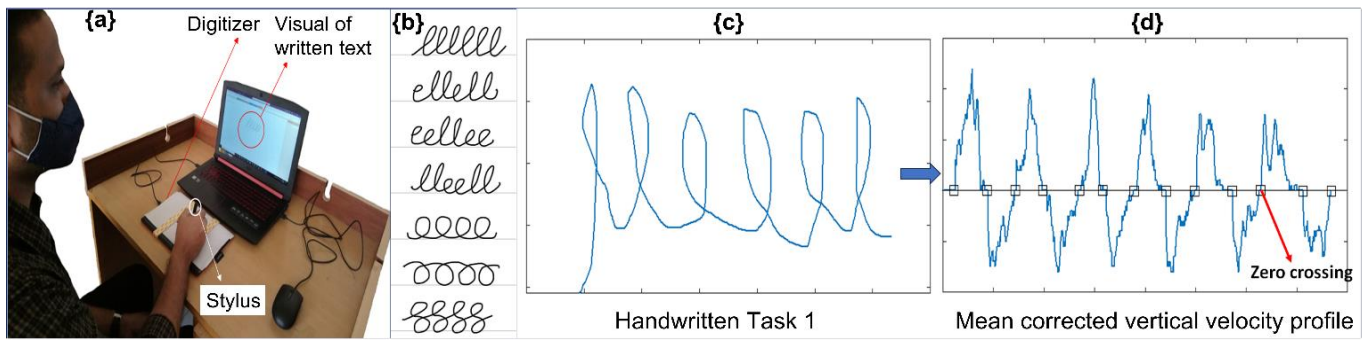

Fig 1. Instrumentation and writing tasks: {a} Experimental Setup, {b} List of seven tasks serially, {c} A sample of handwritten Task 1, {d} Mean corrected vertical velocity profile segmented by zero crossing. Each segment or stroke corresponds to the data available between two zero values in the vertical velocity profile and its corresponding time stamp value. Each such stroke is then used for feature extraction done manually for about 25 relevant features.

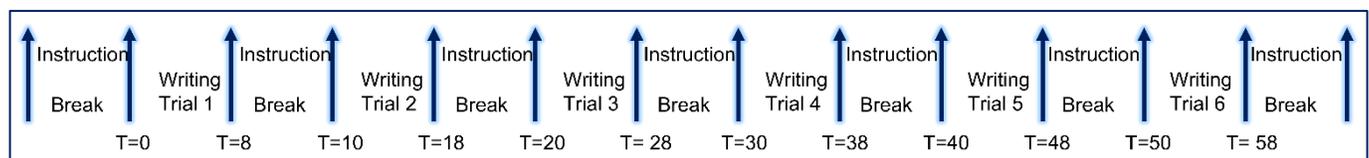

Fig 2. Experimental protocol for one out of seven tasks of experiment, with 6 trials each for dominant as well as non-dominant hand. Same protocol was followed for rest of the tasks. Instructions were given in the beginning of each window.

### C. Instrumentation and Data Acquisition

The experiments were conducted, and all online handwriting data were captured using a commercially available digitizer (WACOM Pen-Tab, CTL-672) which acted as an input device, also shown in Fig. 1 {a}. The spatial data (x and y coordinates) as well as the temporal data (time stamps) were recorded at a sampling rate of 134 Hz. Participants were asked to perform certain writing tasks using their dominant hand with WACOM digitizer, to get familiarized with the experimental setup. All the tasks were performed with both hands, dominant and non-dominant.

### D. Experimental Design and Protocol

The experimental setup and protocol followed in this experimental design is shown in Fig. 1 {a} and Fig. 2 respectively. A set of seven writing tasks of which four non-sensical words in cursive - "*llllll*", "*ellell*", "*eellee*", "*lleell*", three types of symbols - "*QQQ* (4 loops)", "*inverted QQQ* (4 loops)", and "*SSSS*(4 loops)" were used in this experiment as shown in Fig. 1 {b}. Based on handwriting-based tests for handedness [35], Parkinson's [27], etc. the usage of nonsensical words, such as two lowercase cursive (ll)s have been found. Three additional nonsensical words with e's and l's were added to this method (both these letters have great significance in handwriting based assessments done till now) in this study. English and Hindi alphabets, circles, and spirals were investigated during preliminary evaluations. However, the handwriting task analysis was limited to non-sensical phrases and repetitive symbols for the following reasons:

1) Single-digit alphabets showed little inter-hand variability in a statistical examination of handwriting strokes. These jobs were not analyzed due to redundancy, which were bad for grading.

2) Drawing assignments like spirals and circles were purposefully excluded from study due to their difficulty even for dominant hands and redundancy in features.

As per the experimental protocol the participants were made familiarized with the experiment before online handwriting recording. They were asked to practice writing using digitizer on One Note (MS Office product) with their dominant hand. They were made aware of the experiment screen which contained a bounding box inside which they had to record their handwriting within 8 seconds from the time of initiation of recording screen. Before starting each task, instructions were given in the form of visuals on the screen. Each task consisted of two sets, one to be performed with dominant hand and other to be performed with non-dominant hand, each set had 6 trials. In between two trials there was a gap of 2 seconds for screen change which also served as a rest time for participants to avoid fatigue, meanwhile the current task instruction was also displayed on the screen. Participants were instructed to follow their natural handwriting before the beginning of the experiment.

### E. Subject cohort and dataset

There are two important aspects of the data set from a training point of view: the intrasubject data variability and the intersubject variability. Since the total numbers of samples in data matrices were 21030 and 22314 for dominant and non-dominant hand respectively, the average number of samples for a subject was around [43344/43 = 1008]. These 1008 samples were 1008 different strokes which evolved from 7 different combinations of writing tasks. However, the versatility in terms of subjects was not compromised since the chosen subjects belonged to different parts of the degree of handedness spectrum, but their number was restricted to 43, since 43 subjects enrolled for this study; additionally, ambidextrous subjects were difficult to find for this study. This implies that the dataset comprised of information from 86 (43 from left and 43 from right) different distributions. A balanced dataset of left hander and right hander males and females which contained individuals who varied from being ambidextrous to partially unidextrous to completely unidextrous was used. Additionally, the volume of datapoints were sufficient to generate classification accuracies for individuals. Therefore, though the subject number was 43 here the above discussed positives have taken care of the effects of using a small population subset.

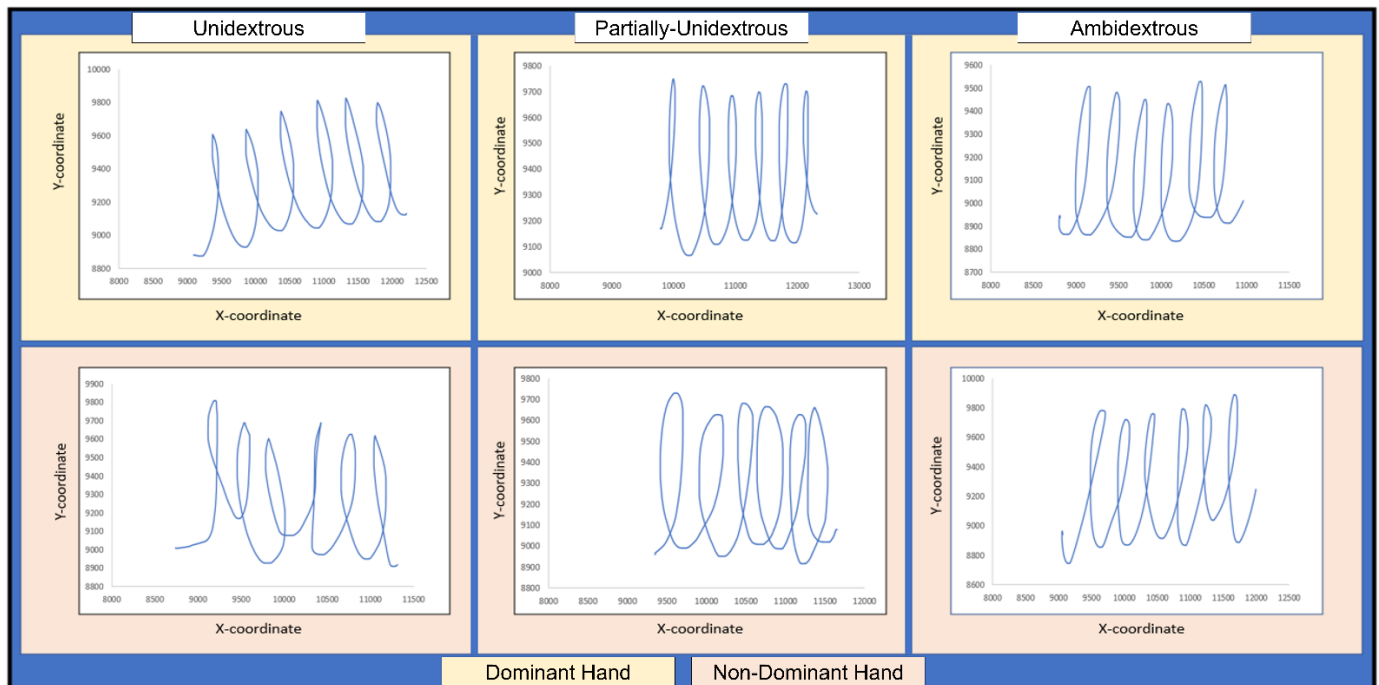

Fig 3. Handwriting samples extracted from groups: "Unidextrous", "Partially-Unidextrous", and "Ambidextrous" for both dominant and non-dominant hand.

TABLE I
EXTRACTED HANDWRITING FEATURES

| Time Features | Static Features | Dynamic Features |
|---|---|---|
| Start Time | Initial Vertical Position | Peak Horizontal Velocity |
| Time Duration | Vertical Size | Peak Horizontal Acceleration |
| Relative Time to Peak VV[#] | Initial Horizontal Position | Peak Vertical Velocity |
| | Horizontal Size | Peak Vertical Acceleration |
| | Straightness Irregularity | Average Absolute Velocity |
| | Slant | Absolute Vertical Jerk |
| | Loop Surface Area | Absolute Jerk |
| | Relative Initial Slant | NPA Points Per Segment[***] |
| | Absolute Size | Average Pen Pressure |
| | Segment length | Number Of Strokes |
| | | Energy |
| | | PV |

[#] Vertical Velocity, [***] Number of Peak Acceleration Points Per Segment/stroke

*F. Feature Extraction and Signal Processing*

Initial pre-processing techniques were applied to filter out any high frequency noise components using a 4th order Butterworth filter at cutoff frequency 15 Hz [36]. After pre-processing vertical velocity was extracted from pre-processed data and was used for segmentation. This method considered one segment to range in between two zero-crossings of vertical velocity [37], which is schematically illustrated in Fig. 1 {c} and {d}. All the segmented parts obtained from one trial were termed as the strokes of that trial from which dynamic and static features were extracted. A co-linearity test was done to remove any redundant features. Therefore, the dataset consisted of thousands of labelled datapoints where each datapoint corresponded to a particular stroke and its 25 features *{6 trials * 7 tasks * 2 hands * 43 subjects * 12 strokes (approximate average) ~= 43344 strokes, from which a dataset of features could be formed, where features formed the columns with 43344 labelled rows}*. Salient handwriting features are listed in Table I and described in supplementary material. The handwriting samples extracted from groups: "Unidextrous", "Partially-Unidextrous", and "Ambidextrous" for both dominant and non-dominant hands are shown in Fig. 3.

*G. Feature Analysis*

From the pre-processed strokes data *(all tasks, all subjects)* handwriting features were extracted, and a dataset was created having two labels- one for dominant hand features "D" and the other for non-dominant hand features "ND". The features in this dataset had different scales and range; hence, each feature in a trial was normalized by the maximum and minimum value obtained in that trial using "MinMax" scaler function of scikit-learn package *(a well-known standardization technique for data scaling which did not affect the nature of distribution but rescaled it in the range of 0 to 1 for easy comparison and analysis). Note: Each writing trial can have many strokes/segments, and each segment can generate 'n' number of features. Standardization of each feature was done by the maximum and minimum value of that feature obtained in a particular trial.*

*H. Assessment and Grading of Degree of Handedness*

*Part A. Statistical method: Davies-Bouldin Index based grading- A model-less approach.*

In this case, it was assumed that there existed only two clusters having their respective centroids. The sum of distances between centroids of these two clusters for all features would not be same for two subject as shown in Fig. 4 {a}. Thus, Davies-Bouldin Index (DB-Index), a well-known method to evaluate the goodness of clustering and measures the inter-cluster centroid distance, was chosen in this work.

Specifically, well segregated clusters show poor DB-Index while mixed clusters show higher DB-Index. Hence, it was hypothesized that "Unidextrous" subjects could show lower values of DB-Index while "Ambidextrous" subjects could show higher values of DB-Index considering that same feature present in dominant and non-dominant hand could be treated as two clusters.

Features of D were denoted as ($f_1D$, $f_2D$, $f_3D$, . . ., $f_kD$) and features of ND as ($f_1ND$, $f_2ND$, $f_3ND$, . . ., $f_kND$), where k meant the best features obtained post feature analysis. D and ND consisted of subject-wise smaller datasets namely $D_{S1}$, $D_{S2}$, $D_{S3}$, . . ., $D_{S43}$ and $ND_{S1}$, $ND_{S2}$, $ND_{S3}$, . . ., $ND_{S43}$ for dominant hand and non-dominant respectively, where S1, S2, …, S43 were subject IDs attached as suffix.

DB index was calculated for 'k' best features, summed together to resultant final DB score, for a particular individual say S1 as represented in (1). Similarly, DB score for each subject was assigned.

$$DB\ score_{(subject\ S1)} = \sum_{i=1}^{k} DB\ score(f_iD_{S1}, f_iND_{S1}) \qquad (1)$$

This method inherently can give the degree of handedness but not the direction i.e., it cannot tell if a person is left-handed or right-handed.

*Part B. Machine learning and deep learning based grading- A model-based approach.*
*1. Machine Learning Algorithms*

Different machine learning methods were explored to measure degree of handedness based on classification accuracies of class separability. Logistic Regression, SVM, Naïve Bayes, KNN, Decision Tree, and Random Forest were implemented. The accuracies were tested with Stratified 10-fold cross-validation using popular scikit-learn python package. In each fold the model was trained again.

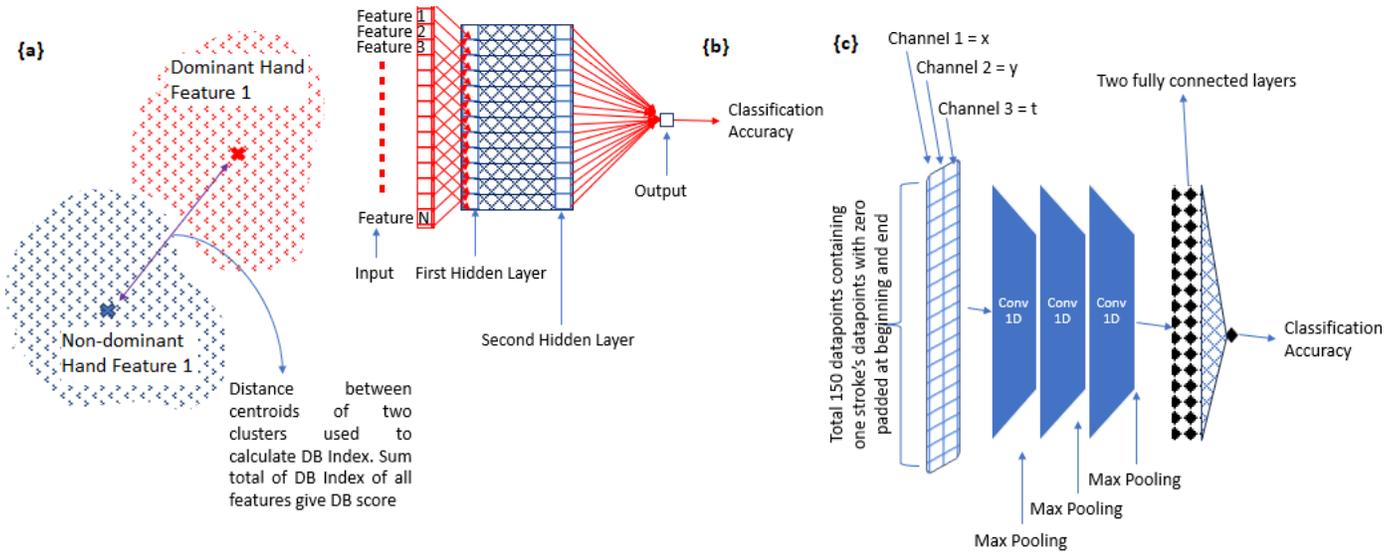

Fig. 4. {a} The illustration shows the DB Index measurement between two centroids of two clusters. {b} The illustration shows the MLP architecture which are fed with features extracted from each stroke. {c} The illustration shows the CNN architecture which was fed with three channel data of each stroke with zero-padding.

The known parameter tuning was applied to the above-mentioned algorithms. The accuracies were tested without feature selection and with feature selection using dedicated feature ranking *(feature ranking explained in supplementary)* methods in each algorithm. Further improvement in the accuracy levels were observed on feature selected category after bagging classifier was added to each conventional algorithm.

*2. A multilayer perceptron model for classification*

A two-layer perceptron model *(neural network)* was trained and tested with Stratified 10-fold cross-validation to check if the accuracy improved in this case. Leave-one subject out method was applied on the two-layer perceptron model to get the accuracy of classification. Feature selection based on feature ranking was applied before feeding the data to the neural network. The network consisted of 2 hidden layers with 12 neurons each with ReLU activation function. Ablation study of MLP is shown in supplementary. Network architecture is shown in Fig. 4 {b}.

*3. A convolutional neural network for classification*

*(a) Ablation study*: From the ablation study done on number of layers, number of filters, number of neurons of fully connected layers, number of layers of fully connected layers (all explained in detail in the supplementary material) it was found that three 1D convolution layers with 128, 64, 32 filters along with two fully connected layers of 20 neurons each yielded the optimum classification accuracy in stratified 10-fold cross validation method. The kernel size of 3 was used. Early stopping for epochs *(maximum epochs were below 100)*, a learning rate of 1e-05, mini-batch size of 32, ReLU activation and Adam optimizer were used heuristically. The network architecture is shown in Fig 4 {c}.

Since the strokes were considered as the input for 1$^{st}$ convolution layer, the original stroke data were padded in the beginning and end with zeroes to reach a uniform range of 150 data values in any stroke. The stroke values contained x, y, and t values which were fed as 3 channels in a 1D convolutional layer. Max-pooling was applied after each convolutional layer along with batch normalization. The output layer consisted of one neuron which yielded classification accuracy of dominant vs non-dominant.

*(b) Training and Testing:* The number of subjects and datapoints pose a big problem when the neural network learns the task-irrelevant features i.e., noise. Evaluating a neural network on training data itself could unveil its underfitting or overfitting characteristics. For grading of individual's DoH leave-one-subject-out method of training was used. This method split the data into tests and training based on population distribution. Each fold used the same initial random weights of the network to generate the same initial random states. For every fold, forty-two individuals' data were split into training (80%) and validation (20%) sets, while one subject's data was for testing set. The validation set prevented overtraining, whereas the testing set evaluated the trained network to provide the classification accuracy for that subject.

*Part C. Grading*

It was hypothesized that for a subject, if the dominant vs non-dominant classification accuracy was found to be very high, then there existed a significant inter-class variability which could be expected in "Unidextrous" category, while for a subject with a poor classification accuracy, it could be said that the classes were mixed or fused together which could be expected in "Ambidextrous" category. Hence, the leave-one-subject-out method of training over k-fold cross validation was preferred, so that the testing accuracy on the left-out subject would be related to his/her degree of handedness. The subject wise classification accuracies generated by MLP and CNN were obtained. These accuracies were converted to a 4-point grading score (0-4) using the following step –

The maximum accuracy for any classification could reach 100 which was mapped to 4 while minimum accuracy which could tend to reach 0 was mapped to 0, with this step the decreasing score from 4 to 0 represented increasing order of "ambidexterity" and hence lower degree of handedness and poor classification accuracy.

### I. Comparison with the Edinburgh Inventory

*1. DB score and EI score comparison*

To represent the accordance of Davies Bouldin score with Edinburgh Inventory score (EI score), Bland Altman plots, correlation, and root mean square error (RMSE) values were obtained. For the Bland Altman plot following steps were followed:

(i) The EI score lies between [-100, 100], where -100 corresponds to pure left-hander and 100 corresponds to pure right-hander, and 0 refers to complete ambidextrous. In our study a pure left-hander or a pure-right hander is "Unidextrous", hence we folded [-100, 100] into [0, 100] by taking absolute of the values, then 0 referred to ambidexterity and 100 referred to "Unidexterity". DB score and EI score were inversely proportional to each other, therefore, EI scores were subtracted from the maximum value i.e., 100, to make them directly proportional to DB scores. EI scores had a different range than DB scores hence, following steps ii, iii, iv were used to bring down EI scores to the range of DB scores.

(ii) EI score as a function of DB score was obtained using curve fitting and was found to be non-linear *(due to the semi-quantitative nature and poor resolution of EI score)* as shown in equation (2).

$$y = -3.1055x^2 + 44.134x - 64.503 = f(x) \quad (2)$$

(iii) The inverse of this function was taken, as shown in equation (3).

$$G(y) = f^{-1}(y) = x \quad (3)$$

(iv) Image of EI score was computed from the inverse function, which was termed as scaled EI. Thus, non-linear scaling was performed.

$$h = G(EI) \quad (4)$$

Where h= scaled EI, and x, y, h $\in R^+$

(v) The correlation and RMSE values were obtained from the scaled EI and DB scores.

*2. Neural networks and EI score comparison*

Similarly, accordance of 4-point scores with EI score was obtained using Bland Altman plots, correlation and RMSE values. EI score as a non-linear function of 4-point score obtained from MLP using curve fitting was found in equation (5). Similarly, EI score as a non-linear function of 4-point score obtained from CNN using curve fitting was found in equation (6).

$$y = 0.2163e^{1.4724x} \quad (5)$$

$$y = 0.0950e^{1.6726x} \quad (6)$$

Above listed steps (i) to (v) were followed for this except the additional subtraction part mentioned in (i) since 4-point scores and EI score were directly proportional inherently.

### J. Statistical Analysis

As per analysis there were different groups: left-handers, right-handers, males, and females. The parametric unpaired t-test and non-parametric Mann-Whitney test was performed depending upon the normality of each group's distribution which was tested by Kolmogorov-Smirnov test. The statistical comparisons were done (a) between males' and females' degree of handedness scores generated by either MLP or CNN, (b) between left and right handers' degree of handedness scores generated by either MLP or CNN.

### K. The role of individual tasks

The goodness of each task in contributing towards degree of handedness assessment was tested using Random Forests with feature selection and bagging classifier using stratified 10-fold cross validation.

## III. RESULTS

### A. Assessment and Grading of Degree of Handedness

*Part A. Statistical method: Davies-Bouldin Index based grading- A model-less approach.*

The DB scores for unidextrous subjects were found to lie in the range [2.33, 4.78], for partially-unidextrous - [4, 7.22] and that for ambidextrous- [7.38, 7.39] as shown in column "DBS" of Table II.

*Part B. Machine learning and deep learning based grading- A model-based approach.*

*1. Machine Learning Algorithms*

The classification accuracies of dominant vs non-dominant hand on the dataset using conventional machine learning models were obtained which is mentioned in Table III. It shows results 'without feature selection', 'with feature selection', and 'with bagging classifier *(added to base classifier)* and feature selection'. All accuracies listed in Table III are the average of the accuracies obtained from Stratified 10-fold cross validation. The best model was made with Random forests with feature selection and bagging classifier.

*2. Multi-layer Perceptron*

A two-layer neural network which was trained and tested for the same dataset had the average accuracy for this classification improved to 86% ± 3.05% for a Stratified 10-fold cross validation. The accuracies outperformed the accuracies obtained from the machine learning algorithms. Additionally, a trend inverse to DB scores was observed in classification accuracies on leave-one-subject-out method with this neural network which is shown in column "MLP Accuracy" of Table II. Neural networks showed high classification accuracy for low DB scores implying higher segregation and higher degree of handedness. Poor classification accuracies and high DB scores implied mixed handed or ambidextrous subjects with lower degree of handedness. These accuracies were converted to a 4-point grading score (0-4) shown in column "Degree of Handedness 4-point score" in Table II. The calculated Edinburgh Inventory scores are also presented in column "EIS" of Table II. The 4-point score corresponded to the degree of handedness just as DB score. Thus, a value closer to 0 was perceived as high ambidexterity and vice versa.

The MLP generated 4-point score for unidextrous subjects were found to lie in the range [3.177, 3.912], for partially unidextrous- [2.578, 3.265] and that for ambidextrous- [1.919, 2.502] as shown in Table II.

*3. Convolutional Neural Network-based Classification*

The average classification accuracies of CNN was found to be 95.06±3.08% under stratified 10-fold cross-validation. The classification accuracies of leave-one-subject-out is shown in

Table II under "CNN Accuracy" and were found to be higher for "Unidextrous" subjects than that of "Partially-Unidextrous" and "Ambidextrous" subjects similar to the observation corresponding to MLP. It also showed the inverse relation between classification accuracy and ambidexterity. The CNN generated 4-point score for unidextrous subjects were found to lie in the range [3.306, 3.930], for partially unidextrous- [2.741, 3.355] and that for ambidextrous- [2.294, 2.658] as shown in Table II.

## B. Comparison with the Edinburgh Inventory using Bland Altman Plot

This analysis dealt with comparing DB scores and 4-point scores $^{(MLP, CNN)}$ with Edinburgh Inventory. Since there existed a non-linear relation of DB score with EI score, therefore, EI score was scaled down non-linearly as discussed in methods for both DB score and 4-point scores. The Bland Altman plot for DB score and scaled EI score is presented in Fig. 5 {a}, that of 4-point score $^{(MLP)}$ and scaled EI score is presented in Fig. 5 {b}, and that of 4-point score $^{(CNN)}$ and scaled EI score is presented in Fig. 5 {c}. It was found that more than 90% *(90.6%)* of the DB scores and networks graded scores lied within 95% confidence interval of the EI scores. The correlation values with scaled EI score were found to be 0.87 with DB score and (0.91 and 0.91) with 4- point scores $^{(MLP\ and\ CNN)}$ respectively. It is interesting to note that the EI scores' resolution were limited, and the resolution could be improved with statistical and deep learning methods. The average RMSE values in percentage turned out to be 7.5% for 4-point scores $^{(MLP\ and\ CNN)}$ and 15.0% for DB score. These results also suggests that assessment between a semi-quantitative and quantitative method should be limited to test of agreement and not direct comparison, since in semi-quantitative method the measurements were done through questionnaires which might have contained personal bias of participants, and the limited resolution of this method is another factor for the same.

TABLE II
COMPARISON OF DEGREE OF HANDEDNESS SCORES OBTAINED FROM EDINBURGH INVENTORY QUESTIONNAIRE, DB SCORES, AND NEURAL NETWORK ACCURACIES

| Sub ID | Class | Subject Characteristics | DBS | MLP Accuracy | 4-point score$^{MLP}$ | CNN Accuracy | 4-point score$^{CNN}$ | EIS |
|---|---|---|---|---|---|---|---|---|
| S35 | U | UT ND arm | 2.47 | 97.81 | 3.912 | 98.25 | 3.93 | -70 |
| S16 | U | UT ND arm | 2.33 | 97.50 | 3.900 | 98.01 | 3.920 | 70 |
| S21 | U | UT ND arm | 2.83 | 97.15 | 3.886 | 97.78 | 3.911 | 70 |
| S39 | U | UT ND arm | 2.89 | 97.10 | 3.884 | 97.69 | 3.908 | -60 |
| S18 | U | UT ND arm | 3.14 | 96.32 | 3.853 | 97.68 | 3.907 | 70 |
| S29 | U | UT ND arm | 3.01 | 94.09 | 3.764 | 97.38 | 3.895 | -50 |
| S33 | U | UT ND arm | 3.41 | 93.47 | 3.739 | 96.69 | 3.868 | -60 |
| S43 | U | UT ND arm | 3.01 | 93.11 | 3.724 | 95.92 | 3.837 | 70 |
| S38 | U | UT ND arm | 3.91 | 93.04 | 3.722 | 95.82 | 3.833 | -60 |
| S20 | U | UT ND arm | 3.39 | 92.25 | 3.690 | 95.50 | 3.82 | 70 |
| S36 | U | UT ND arm | 3.29 | 92.20 | 3.688 | 93.09 | 3.724 | -50 |
| S1 | U | UT ND arm | 3.88 | 90.67 | 3.627 | 92.00 | 3.68 | -50 |
| S3 | U | UT ND arm | 3.52 | 89.92 | 3.597 | 91.63 | 3.665 | -70 |
| S41 | U | UT ND arm | 3.57 | 89.82 | 3.593 | 91.01 | 3.640 | 40 |
| S14 | U | UT ND arm | 2.95 | 89.14 | 3.566 | 90.67 | 3.627 | 30 |
| S40 | U | UT ND arm | 4.07 | 88.71 | 3.548 | 89.53 | 3.581 | 40 |
| S10 | U | UT ND arm | 3.86 | 88.03 | 3.521 | 89.20 | 3.568 | 40 |
| S30 | U | UT ND arm | 4.31 | 86.87 | 3.475 | 88.58 | 3.543 | -30 |
| S19 | U | UT ND arm | 4.20 | 84.87 | 3.395 | 87.01 | 3.480 | 30 |
| S12 | U | UT ND arm | 4.41 | 84.54 | 3.382 | 86.53 | 3.461 | 40 |
| S27 | U | UT ND arm | 4.17 | 84.00 | 3.360 | 85.85 | 3.434 | -30 |
| S42 | U | UT ND arm | 4.78 | 83.42 | 3.337 | 84.87 | 3.395 | 30 |
| S8 | U | UT ND arm | 4.73 | 83.35 | 3.334 | 84.86 | 3.394 | 30 |
| S11 | PU | PT ND arm (S)* | 4.70 | 81.62 | 3.265 | 83.88 | 3.355 | 30 |
| S32 | U | UT ND arm | 4.61 | 81.21 | 3.248 | 83.24 | 3.330 | -30 |
| S26 | PU | Gross motor preferred in ND arm | 4.13 | 80.00 | 3.200 | 82.63 | 3.305 | -30 |
| S23 | U | Gross motor preferred in ND arm | 4.20 | 79.42 | 3.177 | 82.65 | 3.306 | -30 |
| S17 | PU | PT ND arm (S)* | 4.00 | 78.97 | 3.159 | 81.62 | 3.265 | 30 |
| S31 | PU | Gross motor preferred in ND arm (S)* | 5.14 | 78.52 | 3.141 | 80.67 | 3.227 | -30 |
| S6 | PU | PF < 1 hour per week | 5.29 | 76.87 | 3.075 | 79.22 | 3.169 | 10 |
| S37 | PU | PF < 1 hour per week | 4.37 | 76.49 | 3.060 | 78.48 | 3.139 | -10 |
| S28 | PU | PF < 1 hour per week | 5.16 | 76.11 | 3.044 | 78.14 | 3.126 | -10 |
| S24 | PU | PF < 1 hour per week | 5.65 | 74.32 | 2.973 | 77.77 | 3.111 | -10 |
| S4 | PU | PF < 1 hour per week | 5.08 | 72.52 | 2.901 | 75.28 | 3.011 | -10 |
| S5 | PU | PF < 1 hour per week | 5.24 | 70.30 | 2.812 | 74.70 | 2.988 | 10 |
| S9 | PU | PF < 1 hour per week | 5.41 | 69.42 | 2.777 | 72.03 | 2.881 | 10 |
| S15 | PU | PF > 1 hour per week | 6.25 | 67.00 | 2.680 | 71.40 | 2.856 | 10 |
| S34 | PU | PF > 1 hour per week | 6.22 | 65.92 | 2.637 | 71.15 | 2.846 | -10 |
| S7 | PU | PF > 1 hour per week | 6.42 | 65.24 | 2.610 | 69.13 | 2.765 | 10 |
| S2 | PU | PF > 1 hour per week | 7.04 | 65.05 | 2.602 | 68.96 | 2.758 | -10 |
| S13 | PU | PF > 1 hour per week | 7.22 | 64.46 | 2.578 | 68.52 | 2.741 | 10 |
| S25 | A | Ambidextrous | 7.39 | 62.54 | 2.502 | 66.46 | 2.658 | -10 |
| S22 | A | Ambidextrous | 7.38 | 47.98 | 1.919 | 57.35 | 2.294 | -10 |

The column named "Class" contains the groups U, PU, and A which stands for unidextrous, partially unidextrous, ambidextrous. The column named "Subject Characteristics" contains fields such as "UT ND arm" which means untrained non-dominant arm, "PT ND arm (S)*" which means partially trained non-dominant arm due to sports, "Gross motor preferred in ND arm" which means gross motor activities preferred in non-dominant arm, and "PF < 1 hour per week" which means practice frequency less than 1 hour per week. Column "DBS" stands for DB score and "EIS" stands for the Edinburgh Inventory score.

TABLE III
DOMINANT AND NON-DOMINANT HAND CLASSIFICATION ACCURACIES USING CONVENTIONAL MACHINE LEARNING ALGORITHMS

| Machine Learning Algorithms | Parameters | Without Feature Selection Accuracy* | With Feature Selection Accuracy* | With Bagging Classifier & Feature Selection Accuracy* |
|---|---|---|---|---|
| Logistic Regression | L-BFGS optimization, Maximum iterations 100 | 72.88±1.02 | 73.51±0.90 | 73.52±1.09 |
| Decision Tree | Gini used to determine impurity of the split | 71.63±0.88 | 71.64±1.34 | 71.66±1.75 |
| KNN | Brute-force algorithm was used, leaf size 30 | 74.25±0.56 | 74.66±0.11 | 74.62±0.89 |
| Naïve Bayes | No parameters were adjusted, default | 64.12±0.70 | 64.12±0.90 | 64.12±1.05 |
| SVM | Radial basis function kernel with "gamma" | 76.41±0.86 | 79.43±1.34 | 80.02±2.87 |
| Random Forest | Gini impurity, number of trees 100 | 79.6±1.80 | 79.52±1.01 | 82.37±2.05 |

*All reported accuracies correspond to average of stratified 10-fold cross validation accuracies.

## C. Statistical Analysis

Significant differences ($p<<0.05$, unpaired t-test) were observed in gender, where the MLP and CNN generated accuracies of male group differed significantly from that of female. This shows that gender wise there were some significant dissimilarities in the handwriting of the subjects in our study. However, no significant difference ($p=0.6443$, Mann Whitney U test on MLP generated accuracies) and ($p=0.7135$, Mann Whitney U test on CNN generated accuracies) were observed between left handers and right handers

## D. The role of individual tasks

In one of the preliminary analysis, it was found that the tasks affected the classification accuracies. The analysis highlighted the role of tasks on these classification accuracies. The Stratified 10-fold cross validation was applied and the best accuracy along with average accuracy for each task were calculated using Random Forests with feature selection and bagging classifier since it being the best performing machine learning algorithm in this study as shown in Table III. Results of each task's accuracy are shown in Table IV. From Table IV, it was found that Task 1 performed poorly while Task 7 gave the highest mean accuracy for classification revealing its higher differentiating potential in classification.

## IV. DISCUSSION

Despite being an important assessment tool in behavioral, medical, and psychometrical research, the degree of handedness has been a technologically neglected area where the reliance on semiquantitative methods such as questionnaires is still prominent. In this work, we have quantitatively automated the grading of this important aspect of human behavior through simple writing tasks using different computational approaches and their results. Writing has been chosen since it being effective in measuring minute differences and its relation to the brain which is directly measurable. The complexity of handwriting, in fact, appears to be sensitive to cognitive functioning, motor control, and interhemispheric interactions. Studies using functional magnetic resonance imaging (fMRI) suggest that handwriting engages the brain in a more holistic way than typing and any other motor task [38].

Distinction associated with troubled handwriting could be visualized in letters of non-sensical words that demand more complex hand movements such as producing curved trajectories which requires fingers and wrist movements in a certain pattern to produce motion. Therefore, the chosen tasks were capable of revealing significant information.

TABLE IV
TASK EFFECT ON CLASSIFICATION ACCURACY TESTED ON RANDOM FOREST WITH FEATURE SELECTION AND BAGGING CLASSIFIER

| Task Number | Best Accuracy* | Mean Accuracy** |
|---|---|---|
| Task 6 | 84.89 | 81.5 |
| Task 4 | 84.64 | 79.95 |
| Task 7 | 85.04 | 78.39 |
| Task 3 | 80.31 | 77.36 |
| Task 2 | 83.8 | 76.73 |
| Task 5 | 79.91 | 76.18 |
| Task 1 | 69.01 | 62.03 |

*Denotes best accuracy of Stratified 10-fold cross validation and ** denotes average accuracy of the same.

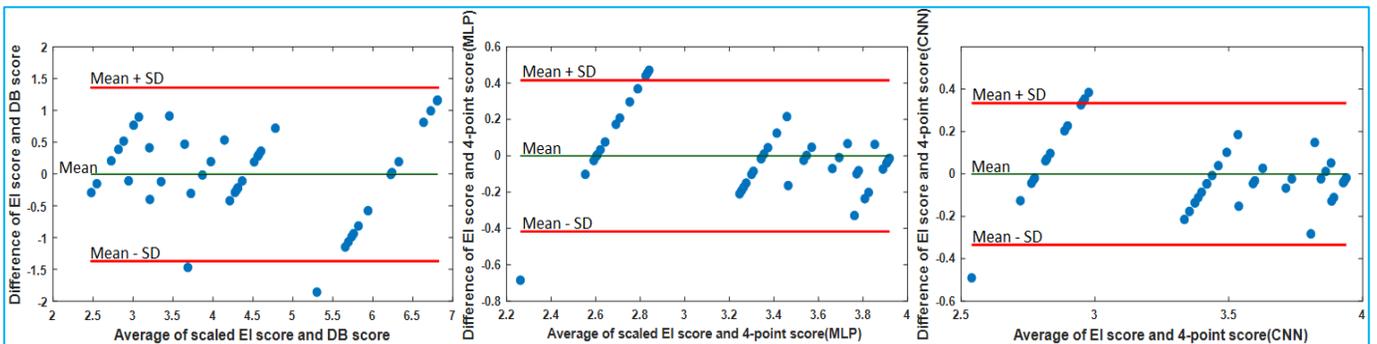

Fig. 5. The figure shows the Bland Altman Plot of EI scores with DB score, 4-point score (MLP), and 4-point score (CNN) respectively.

In this work, the grading of DoH was progressively done through statistical, machine learning, and deep learning methods stage by stage. At each stage there were certain advantages and limitations to the chosen method, discussing which can help picking the appropriate grading technique.

*1) DB score: Pros-* The significant advantage of DB score is that it can assess the degree of handedness with a quantitative value which can give more resolution than any questionnaire, also, it doesn't depend upon the subject number rather the number of features. Thus, it can be used without training and with few electrical resources, making it portable. It is more robust than any verbal assessment.

*Cons-* After a certain limit, manual feature extraction cannot enhance resolution to graded scores. In systems requiring non-linear feature combinations, DB scores may not produce the desired outcomes. The maximum value of DB score depends on the number of features selected for calculations which gives it an obligation to choose and fix the best of features which need to be discriminatory rather than being redundant.

*2) Neural networks based 4-point scores: Pros-* The output generated by neural network gave higher resolution into subject's degree of handedness quantification. It could work well with non-linear relations between the features. In this study while the MLP determined the classification accuracies well with the manually extracted features, the CNN based classifier extracted features automatically and provided better resolutions in degree of handedness scores when compared with the MLP counterpart. Network based classification can capture both degree and direction *(left or right handed)* of handedness efficiently which is an additional gain over DB scores which cannot measure direction.

Here in this study, CNN based classification model outperformed DB Index, machine learning algorithms, and MLP by giving high resolution to individual's classification accuracy. The results of MLP and CNN were, however, comparable which showed that the features fed to MLP were indeed the good features of handwriting strokes.

*Cons-* For an effective model, training data must have intersubject variability, meaning participants should come from different handedness spectrums. Hence, the CNN based 4 point score was an effective method by being quantitative and removing any personal bias.

Furthermore, the occurrence of comparable results within undertaken methods showed the effectiveness of the investigated task in providing resolution to degree of handedness assessment. First-timers had trouble adjusting curves and different altitudes of writing assignments with their non-dominant hand, while horizontal motions and stretching were fine *(feedback from the participants)*. However, task-based classification performances (Table III) may suggest removing tasks with poor classification accuracy from degree of handedness evaluation. Otherwise, these tasks can be training initiators for amputees' and post-stroke survivors' non-dominant hand training due to easy adaptability with tailored strategies.

*Observations:* In this study, degree of handedness was quantified statistically by DB score and by neural network based 4-point scores. It was observed that in any class "U", "PU", or "A," all subjects did receive varied DB scores and neural network accuracies unlike Edinburgh Inventory scores which were same for many subjects under different classes. This implied that the Edinburgh Inventory was only a suggestive assessment of a subject's comfort at executing a job with either hand, and the benefits of using computational methods over questionnaire were substantial. The finding that some unidextrous participants had scores (DB and NN based) similar to or overlapped with partially unidextrous, supported the idea that training partly rather than solely affects capabilities instilled in both hands. It further hinted that these individuals may have had bi-hemispherical brain activity despite untrained non-dominant hand, but this investigation could be only made possible by other techniques such as electroencephalography (EEG). It was also observed that the male and female groups varied significantly ($p<<0.05$, Unpaired t test) in degree of handedness scores graded by CNN and MLP, corroborating past findings showing gender-specific brain areas affect handwriting and thus degree of handedness [39]. There was no significant difference observed between left handers and right handers ($p=0.6443$, Mann Whitney U test, on MLP generated accuracies, and $p=0.7135$, Mann Whitney U test on CNN generated accuracies).

*Applications and future work:* Degree of handedness using handwriting can be considered as a unique behavior which can help in (a) person identification for security and other purposes, (b) determining the viability of forced hand conversion for writing activity, which requires extra efforts for those under socio-cultural pressure (c) in post-stroke and amputee rehabilitation for both dominant and nondominant hand impairments (d) degree of handedness based assessment of complex mental conditions like schizotypy and schizophrenia, and learning disorders like dyslexia (e) human behavioral assessments, (f) sports [40] etc. The degree of handedness was quantitatively evaluated successfully in this study therefore the connection between handwriting output and brain activation levels would be investigated in future studies from the perspective of degree of handedness. Further, this study will be implemented on a large population cohort to evaluate the changes in the resolution of each subject's score. The assessment and gaining insight into degree of handedness can solve multidimensional problems concerned with neuroscience, rehabilitation, physiology, psychometry, behavioral sciences, and forensics.

*Limitations:* The study could not find many ambidextrous subjects which is a limitation. The sample size for this pilot study was kept limited to 43, hence, the performance of neural networks on a larger population size may yield changes in scores, which is a limitation of present work. Due to the small population tested, this study restrained from commenting on significantly different handwriting features of left-handers and right-handers. The presented study is based on handwriting which is a fine motor skill. This study did not evaluate gross motor skill quantitatively which could be a separate study.

## V. CONCLUSION

The presented study dealt with studying handwriting signal in subjects who varied from being "Unidextrous", "Partially-Unidextrous" and "Ambidextrous". In this study, the segmented handwriting signal called strokes were used to quantify degree of handedness using statistical method- DB Index and deep learning methods- MLP and CNN. Traditional methods such as the Edinburgh Inventory questionnaire are semi-quantitative in nature and have less sensitivity to minor variations in dexterity. While the proposed methods are quantitative and continuous in nature with better sensitivity towards dexterity variations. The CNN based automated grading of degree of handedness was found to be a suitable assessment tool. This study also suggested that quantified degree of handedness could be used for multi-faceted use. The proposed methods will be modified in future studies to incorporate activities in addition to writing which will be quantified to generate a universal degree of handedness measure for dexterous as well as gross motor skills (non-dexterous) tasks.

## VI. REFERENCES

[1] J. Fagard, "The nature and nurture of human infant hand preference," *Ann N Y Acad Sci*, vol. 1288, no. 1, pp. 114–123, 2013, doi: 10.1111/nyas.12051.

[2] R. E. Propper *et al.*, "A combined fMRI and DTI examination of functional language lateralization and arcuate fasciculus structure: Effects of degree versus direction of hand preference," *Brain Cogn*, vol. 73, no. 2, pp. 85–92, Jul. 2010, doi: 10.1016/j.bandc.2010.03.004.

[3] C. G. F. de Kovel, A. Carrión-Castillo, and C. Francks, "A large-scale population study of early life factors influencing left-handedness," *Sci Rep*, vol. 9, no. 1, Dec. 2019, doi: 10.1038/s41598-018-37423-8.

[4] J. Martineau, F. Andersson, C. Barthélémy, J. P. Cottier, and C. Destrieux, "Atypical activation of the mirror neuron system during perception of hand motion in autism," *Brain Res*, vol. 1320, pp. 168–175, Mar. 2010, doi: 10.1016/j.brainres.2010.01.035.

[5] D. V. M. Bishop, "Cerebral asymmetry and language development: Cause, correlate, or consequence?," *Science*, vol. 340, no. 6138. American Association for the Advancement of Science, 2013. doi: 10.1126/science.1230531.

[6] A. K. Lindell and K. Hudry, "Atypicalities in cortical structure, handedness, and functional lateralization for language in autism spectrum disorders," *Neuropsychology Review*, vol. 23, no. 3. pp. 257–270, Sep. 2013. doi: 10.1007/s11065-013-9234-5.

[7] M. Lenzenweger, "Schizotypy, schizotypic psychopathology and schizophrenia," *World Psychiatry*, vol. 17, pp. 25–26, Jul. 2018, doi: 10.1002/wps.20479.

[8] H. C. Tsuang, W. J. Chen, S. Y. Kuo, and P. C. Hsiao, "Handedness and schizotypy: The potential effect of changing the writing-hand," *Psychiatry Res*, vol. 242, pp. 198–203, Aug. 2016, doi: 10.1016/j.psychres.2016.04.123.

[9] K. L. Isaacs, W. B. Barr, A.; Peter, K. Nelson, and O. Devinsky, "Degree of handedness and cerebral dominance," 2006. [Online]. Available: www.neurology.org

[10] D. Zapała *et al.*, "The effects of handedness on sensorimotor rhythm desynchronization and motor-imagery BCI control," *Sci Rep*, vol. 10, no. 1, Dec. 2020, doi: 10.1038/s41598-020-59222-w.

[11] E. V. Bobrova *et al.*, "Success of hand movement imagination depends on personality traits, brain asymmetry, and degree of handedness," *Brain Sci*, vol. 11, no. 7, Jul. 2021, doi: 10.3390/brainsci11070853.

[12] R. L. Sainburg and S. V. Duff, "Does motor lateralization have implications for stroke rehabilitation?," *Journal of Rehabilitation Research and Development*, vol. 43, no. 3. pp. 311–322, 2006. doi: 10.1682/JRRD.2005.01.0013.

[13] K. B. Lyle and M. C. Grillo, "Why are consistently-handed individuals more authoritarian? The role of need for cognitive closure," *Laterality*, vol. 25, no. 4, pp. 490–510, Jul. 2020, doi: 10.1080/1357650X.2020.1765791.

[14] S. Christman, "Individual differences in personality as a function of degree of handedness: Consistent-handers are less sensation seeking, more authoritarian, and more sensitive to disgust," *Laterality*, vol. 19, no. 3, pp. 354–367, May 2014, doi: 10.1080/1357650X.2013.838962.

[15] K. B. Lyle, L. K. Chapman, and J. M. Hatton, "Is handedness related to anxiety? New answers to an old question," *Laterality*, vol. 18, no. 5, pp. 520–535, 2013, doi: 10.1080/1357650X.2012.720259.

[16] E. Prichard, R. E. Propper, and S. D. Christman, "Degree of handedness, but not direction, is a systematic predictor of cognitive performance," *Frontiers in Psychology*, vol. 4, no. JAN. 2013. doi: 10.3389/fpsyg.2013.00009.

[17] I. C. McManus, G. Buckens, N. Harris, A. Flint, H. L. A. Ng, and F. Vovou, "Faking handedness: Individual differences in ability to fake handedness, social cognitions of the handedness of others, and a forensic application using Bayes' theorem," *Laterality*, vol. 23, no. 1, pp. 67–100, Jan. 2018, doi: 10.1080/1357650X.2017.1315430.

[18] R. C. Oldfield, "THE ASSESSMENT AND ANALYSIS OF HANDEDNESS: THE EDINBURGH INVENTORY," Pergamon Press, 1971.

[19] H. R. Siebner *et al.*, "Long-Term Consequences of Switching Handedness: A Positron Emission Tomography Study on Handwriting in 'Converted' Left-Handers," 2002.

[20] I. Papousek and G. Schulter, "EEG CORRELATES O F BEHAVIOURAL LATERALITY: RIGHT-HANDEDNESS '," 1999.

[21] I. Lawson, "Purdue Pegboard Test," *Occupational Medicine*, vol. 69, no. 5. Oxford University Press, pp. 376–377, Aug. 22, 2019. doi: 10.1093/occmed/kqz044.

[22] K. A. Provins and P. Cunliffe, "The reliability of some motor performance tests of handedness," *Neuropsychologia*, vol. 10, no. 2, pp. 199–206, 1972.

[23] Y. Bay Ayzeren, M. Erbilek, and E. Celebi, "Emotional State Prediction from Online Handwriting and Signature Biometrics," *IEEE Access*, vol. 7, pp. 164759–164774, 2019, doi: 10.1109/ACCESS.2019.2952313.

[24] L. Likforman-Sulem, A. Esposito, M. Faundez-Zanuy, S. Clemencon, and G. Cordasco, "EMOTHAW: A Novel Database for Emotional State Recognition from Handwriting and Drawing," *IEEE Trans Hum Mach Syst*, vol. 47, no. 2, pp. 273–284, Apr. 2017, doi: 10.1109/THMS.2016.2635441.

[25] C. Kahindo, M. A. El-Yacoubi, S. Garcia-Salicetti, A. S. Rigaud, and V. Cristancho-Lacroix, "Characterizing early-stage Alzheimer through spatiotemporal dynamics of handwriting," *IEEE Signal Process Lett*, vol. 25, no. 8, pp. 1136–1140, Aug. 2018, doi: 10.1109/LSP.2018.2794500.

[26] N. D. Cilia, T. D'Alessandro, C. De Stefano, F. Fontanella, and M. Molinara, "From Online Handwriting to Synthetic Images for Alzheimer's Disease Detection Using a Deep Transfer Learning Approach," *IEEE J Biomed Health Inform*, vol. 25, no. 12, pp. 4243–4254, Dec. 2021, doi: 10.1109/JBHI.2021.3101982.

[27] P. Drotár, J. Mekyska, I. Rektorová, L. Masarová, Z. Smékal, and M. Faundez-Zanuy, "Decision support framework for Parkinson's disease based on novel handwriting markers," *IEEE Transactions on Neural Systems and Rehabilitation Engineering*, vol. 23, no. 3, pp. 508–516, May 2015, doi: 10.1109/TNSRE.2014.2359997.

[28] N. Zhi, B. K. Jaeger, A. Gouldstone, R. Sipahi, and S. Frank, "Toward Monitoring Parkinson's Through Analysis of Static Handwriting Samples: A Quantitative Analytical Framework," *IEEE J Biomed Health Inform*, vol. 21, no. 2, pp. 488–495, Mar. 2017, doi: 10.1109/JBHI.2016.2518858.

[29] D. Impedovo and G. Pirlo, "Dynamic Handwriting Analysis for the Assessment of Neurodegenerative Diseases: A Pattern Recognition Perspective," *IEEE Rev Biomed Eng*, vol. 12, pp. 209–220, May 2018, doi: 10.1109/RBME.2018.2840679.

[30] T. Asselborn, M. Chapatte, and P. Dillenbourg, "Extending the Spectrum of Dysgraphia: A Data Driven Strategy to Estimate Handwriting Quality," *Sci Rep*, vol. 10, no. 1, Dec. 2020, doi: 10.1038/s41598-020-60011-8.

[31] T. Asselborn *et al.*, "Automated human-level diagnosis of dysgraphia using a consumer tablet," *NPJ Digit Med*, vol. 1, no. 1, Dec. 2018, doi: 10.1038/s41746-018-0049-x.

[32] K. P. Feder and A. Majnemer, "Children's Handwriting Evaluation Tools and Their Psychometric Properties," *Phys Occup Ther Pediatr*, vol. 23, no. 3, pp. 65–84, Jan. 2003, doi: 10.1080/j006v23n03_05.

[33] R. Plamondon, C. O'Reilly, J. Galbally, A. Almaksour, and É. Anquetil, "Recent developments in the study of rapid human movements with the kinematic theory: Applications to handwriting and signature synthesis," *Pattern Recognit Lett*, vol. 35, no. 1, pp. 225–235, Jan. 2014, doi: 10.1016/J.PATREC.2012.06.004.

[34]  M. ANNETT, "A CLASSIFICATION OF HAND PREFERENCE BY ASSOCIATION ANALYSIS," *British Journal of Psychology*, vol. 61, no. 3, pp. 303–321, 1970, doi: 10.1111/j.2044-8295.1970.tb01248.x.

[35]  S. Klöppel, A. Vongerichten, T. Van Eimeren, R. S. J. Frackowiak, and H. R. Siebner, "Can left-handedness be switched? Insights from an early switch of handwriting," *Journal of Neuroscience*, vol. 27, no. 29, pp. 7847–7853, Jul. 2007, doi: 10.1523/JNEUROSCI.1299-07.2007.

[36]  H.-L. Teulings and F. J. Maarse, "Human Movement Science 3 (1984) 193-217 North-Holland DIGITAL RECORDING AND PROCESSING OF HANDWRITING MOVEMENTS."

[37]  J. M. Hollerbach, "Biological Cybernetics An Oscillation Theory of Handwriting," 1981.

[38]  M. Karimpoor, N. W. Churchill, F. Tam, C. E. Fischer, T. A. Schweizer, and S. J. Graham, "Functional MRI of handwriting tasks: A study of healthy young adults interacting with a novel touch-sensitive tablet," *Front Hum Neurosci*, vol. 12, Feb. 2018, doi: 10.3389/fnhum.2018.00030.

[39]  Y. Yang *et al.*, "Men and women differ in the neural basis of handwriting," *Hum Brain Mapp*, vol. 41, no. 10, pp. 2642–2655, Jul. 2020, doi: 10.1002/hbm.24968.

[40]  A. J. Marcori, P. H. M. Monteiro, and V. H. A. Okazaki, "Changing handedness: What can we learn from preference shift studies?," *Neuroscience and Biobehavioral Reviews*, vol. 107. Elsevier Ltd, pp. 313–319, Dec. 01, 2019. doi: 10.1016/j.neubiorev.2019.09.019.

## Supplementary Material

Feature Description

***Segment***: A trial of a word was divided into strokes or segments. Segmentation of the word based on zero vertical velocity was performed. The signal portion consisted between two zero vertical velocities was marked as a segment.
***Start Time***: The starting time instant for first stroke was considered as Start Time.
***Time Duration***: Time interval spent during the first and last sample in a stroke is the duration. A very prominent feature which can be well exploited in forgery detection cases, since imitation is often printing, but in the online mode the duration spent by a writer to imitate a word varies greatly from person to person.
***Start Vertical Position***: The position taken by the writer to start the word experiment in vertical direction relative to the horizontal edge of the recording area of the tablet/digitizer. Although a minimally contributing feature but tends to affect the classification when personal behavioural traits were examined.
***Vertical Size***: The distance between $y_{min}$ to $y_{max}$ in a stroke.
***Peak Vertical Velocity***: The maximum vertical velocity in a stroke.
***Peak Vertical Acceleration***: The maximum vertical acceleration in a stroke.
***Start Horizontal Position***: The position taken by the writer to start the word experiment in vertical direction relative to the vertical edge of the recording area of the tablet/digitizer.
***Horizontal Size***: The distance between $y_{min}$ to $y_{max}$ in a stroke.
***Peak Horizontal Velocity***: The maximum horizontal velocity in a stroke.
***Peak Horizontal Acceleration***: The maximum vertical acceleration in a stroke.
***Straightness Irregularity***: Straightness Irregularity is the measure of the deviation of the points from a straight fit, normalized by length of the straight line.
***Slant***: Direction from start point to end point in radians.
***Loop Surface Area***: Area enclosed between two strokes or segments.
***Relative Initial Slant***: Ratio of the initial slant (time restricted to 80ms) to the entire stroke slant.
***Relative Time To Peak Vertical Velocity***: Ratio of the time required to achieve maximum peak velocity (from start time) to the total duration for a segment/stroke.
***Absolute Size***: Absolute size of a stroke/segment.
***Average Absolute Velocity***: Average absolute velocity of a stroke or segment.
***Absolute Jerk***: The Root Mean Square (RMS) value of absolute jerk across all samples of a stroke or segment.
***Number of peak acceleration points***: Number of acceleration peaks in a stroke.
***Average pen pressure***: Average of pen pressure values per stroke/ segment.
***PV***: The product of average absolute velocity and average pen pressure per segment.
***Energy***: Taeger-Kaiser energy associated with peak vertical velocity.

**Feature ranking methods**
Correlation is used for feature selection because favourable variables correlate well with the target. Variables should be uncorrelated yet correlated with the aim. Since two linked characteristics don't provide information, the model just needs one. Hence Spearman's Correlation was utilized. Further, individual features were tested for accuracy using Decision Trees to get the feature ranking.

**Ablation study MLP**
Stratified 10-fold cross validation optimised the network using Task 1 data. Then, the first hidden layer's neuron count was adjusted from 6 to 18 using a 3-neuron step size. The first hidden layer's minimal error neurons were fixed (classification accuracies determined this). Early stopping was used on number of epochs. In the second scenario, a two-hidden layer ANN model was created by varying the number of neurons in the second hidden layer from 6 to 18 with a step size of 3 neurons while fixing the first hidden layer. In training the network for more than two hidden layers, vanishing error gradients were the main concerns. For feed-forward neural network optimization, just two hidden layers were used.

**Ablation study CNN**

The CNN based classification architecture was made by varying the number of layers and filters of convolution layer as well as number and neurons of dense layers. Certain parameters such as epochs, learning rate, optimizer, activation function was heuristically taken which are not shown in this part of the ablation study.

TABLE I

CLASSIFICATION RESULTS FOR DIFFERENT NUMBERS OF CONVOLUTION LAYERS IN THE PROPOSED CNN ARCHITECTURE FOR STRATIFIED 10-FOLD CV. THE $1^{ST}$, $2^{ND}$, AND $3^{RD}$ LAYERS CONTAIN 128, 64, AND 32 FILTERS RESPECTIVELY.

| LAYERS AND FILTERS | ACCURACY | PRECISION |
| --- | --- | --- |
| 1, 128 | 89.99±2.36 | 81.09±1.66 |
| 2, 64 | 92.12±1.16 | 85.88±2.67 |
| 3, 32 | 95.06±3.08 | 89.77±2.80 |

TABLE II

CLASSIFICATION RESULTS FOR DIFFERENT NUMBERS OF DENSE LAYERS IN THE PROPOSED CNN ARCHITECTURE FOR STRATIFIED 10-FOLD CV. THE $1^{ST}$ AND $2^{ND}$ LAYERS CONTAIN 20 NEURONS EACH.

| LAYERS | ACCURACY | PRECISION |
| --- | --- | --- |
| 1 | 91.04±4.13 | 86.23±2.89 |
| 2 | 95.06±3.08 | 89.77±2.80 |

TABLE III

CLASSIFICATION RESULTS FOR DIFFERENT NUMBERS OF NEURONS IN DENSE LAYERS OF THE PROPOSED CNN ARCHITECTURE FOR STRATIFIED 10-FOLD CV.

| NEURONS | ACCURACY | PRECISION |
| --- | --- | --- |
| 10 | 79.02±6.15 | 71.22±3.99 |
| 15 | 87±3.14 | 81.34±5.52 |
| 20 | 95.06±3.08 | 89.77±2.80 |